\input{psfig}

\documentstyle[12pt]{article}
\begin{document}

\hbadness=10000
\setcounter{page}{1}

\title{\vspace*{-0.5cm} \hspace*{\fill}{\normalsize LA-UR-00-309}
\\[1.5ex] 
{\bf A Parallel Algorithm for Dilated Contour Extraction from Bilevel Images
}}

\author{ 
B.R. Schlei\thanks{E. Mail: schlei@LANL.gov}\\[0.3ex] 
{\it  Theoretical Division, T-1,
Los Alamos National Laboratory}\\[3.0ex]
L. Prasad\thanks{E. Mail: prasad@LANL.gov}\\[0.3ex] 
{\it  Nonproliferation and International Security, NIS-7,
Los Alamos National Laboratory\\[4.5ex]} 
} 
\author{
B.R. Schlei${}^1$\thanks{E. Mail: schlei@LANL.gov}{\ } and  
L. Prasad${}^2$\thanks{E. Mail: prasad@LANL.gov}{\ }\\[1.5ex]
{\it ${}^1$MS E541, T-1, Los Alamos National Laboratory, 
Los Alamos, NM 87545}\\
{\it ${}^2$MS E541, NIS-7, Los Alamos National Laboratory, 
Los Alamos, NM 87545}
}

\date{January 12, 2000}

\maketitle

\begin{abstract}
We describe a simple, but efficient algorithm
for the generation of dilated contours from bilevel images.
The initial part of the contour extraction is explained to be a good 
candidate for parallel computer code generation. The remainder
of the algorithm is of linear nature.
\end{abstract}   

\newpage

The notion of {\it shape} is intimately related to the notion of 
{\it boundary} or {\it contour}, in that the contour delineates the 
shape from its exterior, and conversely the shape is defined in extent 
by its contour.\\
In digital image processing contour extraction is important for the 
purpose of enclosing and characterizing a set of pixels belonging to
an object within a given image. 
In the following, we shall work only with bilevel
images, i.e., images that contain only pixels of two spectral
values, e.g., black and white pixels. Without loss of 
generality let us assume that we have therefore only
black and white pixels in a given image, where the white pixels
represent pixels of objects, and black pixels represent pixels
of holes and the universe, respectively. Fig. 1.a shows an example 
of a bilevel image.

\begin{figure} [h]
\begin{center}\mbox{ }
\vspace*{-0.5cm}
\[
\psfig{figure=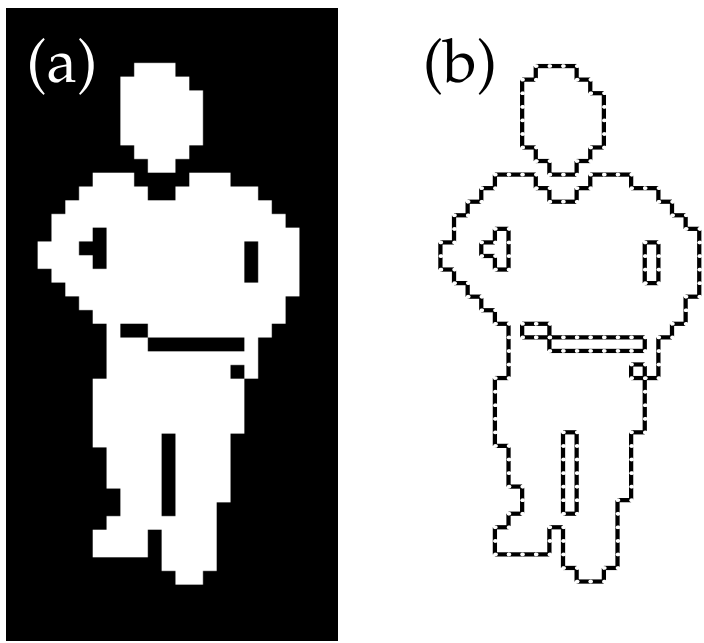,bbllx=0.0cm,bblly=5.5cm,%
bburx=22.0cm,bbury=22.5cm,width=12.0cm,clip=}
\]
\caption{(a) Bilevel image of a person. The shape consists of two
   blobs of which one of them contains a few holes. (b) Contour set 
   of edges, which separates white pixels from black ones. We note, 
   that the edges are not connected yet at this stage of the contour 
   extraction.
} 
\end{center}
\label{fg:fig01} 
\end{figure}

As a first step, we shall construct a set of edges, which 
separates white pixels from black ones. In order to do this,
we note that a given white pixel $i$ has four corners $A^i,B^i,C^i,D^i$
({\it cf.} Fig. 2).
From the pixels corner points one can then construct its four
edges $A^iB^i,B^iC^i,C^iD^i,D^iA^i$. Let $A^iB^i \equiv e_1^i$,
$B^iC^i \equiv e_2^i$, $C^iD^i \equiv e_3^i$ and $D^iA^i \equiv e_4^i$, 
respectively. Then we can always tell for a given edge, $e_j^i$ $(j=1,2,3,4)$, 
the side where the parent pixel is present. Thus, we always know
for a given edge, on which sides the inside and the outside of the
corresponding shape is located.

\begin{figure} [h]
\begin{center}\mbox{ }
\vspace*{-1.5cm}
\[
\psfig{figure=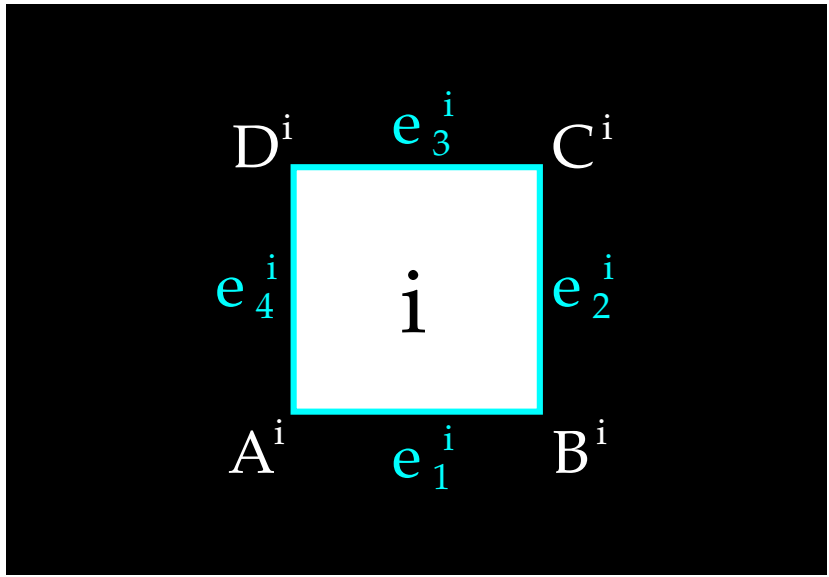,bbllx=0.0cm,bblly=7.0cm,%
bburx=22.0cm,bbury=22.5cm,width=12.0cm,clip=}
\]
\caption{Corners and edges of the $i$-th pixel (see text for more detail).
} 
\end{center}
\label{fg:fig02} 
\end{figure}

A pixel $i$ can contribute to the set of edges, which separates white 
pixels from black ones, either with no, one, two, three or four edges.
If two neighboring pixels are white, the common edge they share, i.e.,
$e_1^i = e_3^{\,j}$ or $e_2^i = e_4^{\,j}$ ($i \neq j$), will {\it not} 
contribute to our edge list. Hence, our desired edge list will only 
consist of edges which have a multiplicity of one. For the 
bilevel image in Fig. 1.a we show the corresponding (but not yet 
connected) edge set in Fig. 1.b.

We would like to emphasize, that all given white pixels could be treated
simultaneously, when their lower, left, upper, and right edges are examined
for candidates in the contour pixel edge set. Thus, here, an excellent 
opportunity for the generation of parallel computer code is given by our 
above described technique.\\

In the second step, we construct connected loops from our contour edge
set which we have generated so far. In doing so, we first enumerate the edges
of the contour edge set. Then, a new set is created, which is the set of
end points (i.e., their coordinate pairs) of the edges themselves. 
Every point is listed only once, but for each point we also list which 
edges (their identities are given through their enumeration) are connected 
to a given point. The reader can convince himself easily, that 
there are always either two or four edges connected to a point
in the point list.

At this stage, all the knowledge for connecting the edges is
available. To be specific, we shall generate contours, which are oriented
counterclockwise about the objects they enclose, and which are oriented 
clockwise when they enclose holes within the objects, respectively.

Initially, all edges of the contour edge set are labeled as ``unused''.
We take the first element of the contour edge set to construct our
first contour loop. If the edge is of type $e_1^i$ ($e_2^i$ ($e_3^i$ 
($e_4^i$))) we note point $A^i$ ($B^i$ ($C^i$ ($D^i$))) as our
first point of the first loop in order to ensure its {\it correct}
orientation. We label our first edge as ``used'' and then look up 
the end point list to examine to which other 
edge our current edge is connected at the other point $B^i$ ($C^i$ ($D^i$ 
($A^i$))) of our used edge. At this point, we have to distinguish
between two cases. In the first case, only two edges are connected
to point $B^i$ ($C^i$ ($D^i$ ($A^i$))), wherein we just consider the
as yet unused edge as our next edge. In the second case,
four edges are connected to point $B^i$ ($C^i$ ($D^i$ ($A^i$))).
Then we choose as the next edge the unused edge belonging to the
current pixel $i$. In doing so, we always ensure a unique choice for 
building the contours. Furthermore,
we weaken the connectivity between pixels, that touch each other only
in one point. In fact, as we shall see below, our dilated versions of 
the contours will lead to a total separation between two pixels that only
share {\it one} common point. 
After having found our next edge $j$, we shall add its first point
to our contour point chain list, and then we label that new edge as ``used''. 
We reiterate the above described procedure, until we encounter our first 
``used'' element of the contour edge set. 
In order to avoid double accounting, we 
shall {\it not} add the first point of the first edge to the contour point
chain again. This, then, concludes the construction of the first contour. 
While creating the contour, we may count the number of edges, which we have 
used so far and compare it to the total number of edges present in our 
contour edge set.

In case there are still ``unused'' edges present in our contour 
edge set we shall scroll through our edge list, until we find the
first next ``unused'' edge and use it as the first edge of our next contour.

We repeat the above described algorithm to create the new contour.
This is done until eventually all edges have been used. The result will 
be a set of point chains, of which each one of them represents a closed
contour.

At this point, we may note that the smallest contours consist of only four 
edge points. The latter result is valid for all isolated white pixels, i.e., 
those, which have no other white pixels as 4-neighbors.\\

Fig. 3.a shows the contours for the bilevel image given in Fig. 1.a. 
We stress, that in our set of point chains each point chain is a list
of the first points of our pixel edges. If we replace these points
by the middle points of the edges without changing their connectivity, 
we get modified contours, which are a dilation of the centers of the 
boundary pixels. The dilated contour version derived 
from the example given in Fig. 3.a is shown in Fig. 3.b. 
In particular, in places where some of the contours touch themselves 
due to two white neighboring pixels, which are in contact in only one point, 
we obtain a definite separation of the contour sections.\\

\begin{figure} [h]
\begin{center}\mbox{ }
\vspace*{-0.5cm}
\[
\psfig{figure=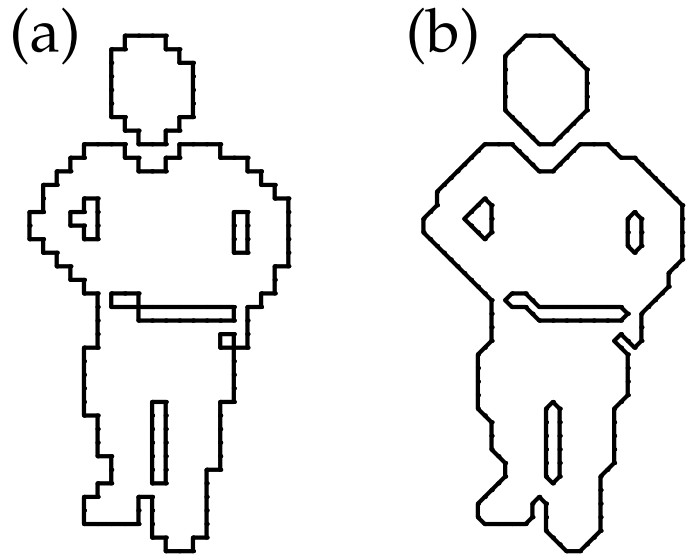,bbllx=0.0cm,bblly=5.5cm,%
bburx=22.0cm,bbury=22.5cm,width=12.0cm,clip=}
\]
\caption{(a) Connected contour set of edges, which separates white pixels 
   from black ones in Fig. 1.a. (b) Dilated contours.
} 
\end{center}
\label{fg:fig03} 
\end{figure}

We would like to mention, that another very popular contour extraction
algorithm is given by Ref. \cite{pavlidis}. But this algorithm is very 
slow compared to our algorithm, since it explores the 8-neighborhood of 
the image pixels in a {\it sequential} raster scan, whereas we use only 
the 4-neighborhood of the pixels and all pixels at the same time. 
Furthermore, we have convinced ourselves that the algorithm presented in Ref. 
\cite{pavlidis} cannot account for nested regions in the image very easily
and reliably.\\

In summary, we have presented a partially parallel and otherwise linear
algorithm to construct dilated contours from bilevel images. We stress,
that the generated contours are always nondegenerate, i.e., 
they always enclose an area larger than zero, and they never cross or 
overlap each other.
Furthermore, our contours are oriented with respect to the objects and their
possible holes.

\end{document}